\documentclass[conference]{IEEEtran}
\IEEEoverridecommandlockouts
\usepackage{cite}
\usepackage{amsmath,amssymb,amsfonts}
\usepackage{algorithmic}
\usepackage{graphicx}
\usepackage{textcomp}
\usepackage{xcolor}
\def\BibTeX{{\rm B\kern-.05em{\sc i\kern-.025em b}\kern-.08em
    T\kern-.1667em\lower.7ex\hbox{E}\kern-.125emX}}
\begin{document}
\title{DynaRAG: Bridging Static and Dynamic Knowledge in Retrieval-Augmented Generation}
\author{
\IEEEauthorblockN{1\textsuperscript{st} Penghao Liang*}
\IEEEauthorblockA{
\textit{Northeastern University} \\
Boston, USA, 02115 \\
liang.p@northeastern.edu
}
\and
\IEEEauthorblockN{2\textsuperscript{nd} Mengwei Yuan}
\IEEEauthorblockA{
\textit{Independent Researcher} \\
Milpitas, USA, 95035 \\
yuanmw1998@gmail.com
}
\and
\IEEEauthorblockN{3\textsuperscript{rd} Jianan Liu}
\IEEEauthorblockA{
\textit{Independent Researcher} \\
Austin, USA, 78613 \\
jiananliu2408@gmail.com
}
\and
\IEEEauthorblockN{4\textsuperscript{th} Jing Yang}
\IEEEauthorblockA{
\textit{Washington University in St. Louis} \\
St. Louis, USA, 63130 \\
jing.y@wustl.edu
}
\and
\IEEEauthorblockN{5\textsuperscript{th} Xianyou Li}
\IEEEauthorblockA{
\textit{New York University} \\
New York, USA, 10012 \\
xl4230@nyu.edu
}
\and
\IEEEauthorblockN{6\textsuperscript{th} Weiran Yan}
\IEEEauthorblockA{
\textit{Independent Researcher} \\
Santa Clara, USA, 95051 \\
yanwr2016@gmail.com
}
\and
\IEEEauthorblockN{7\textsuperscript{th} Yichao Wu}
\IEEEauthorblockA{
\textit{Northeastern University} \\
Boston, USA, 02115 \\
wu.yicha@northeastern.edu
}
}
\maketitle
\begin{abstract}
We present DynaRAG, a retrieval-augmented generation (RAG) framework designed to handle both static and time-sensitive information needs through dynamic knowledge integration. Unlike traditional RAG pipelines that rely solely on static corpora, DynaRAG selectively invokes external APIs when retrieved documents are insufficient for answering a query. The system employs an LLM-based reranker to assess document relevance, a sufficiency classifier to determine when fallback is necessary, and Gorilla v2---a state-of-the-art API calling model---for accurate tool invocation. We further enhance robustness by incorporating schema filtering via FAISS to guide API selection. Evaluations on the CRAG benchmark demonstrate that DynaRAG significantly improves accuracy on dynamic questions, while also reducing hallucinations. Our results highlight the importance of dynamic-aware routing and selective tool use in building reliable, real-world question-answering systems.
\end{abstract}
\section{Introduction}
Large Language Models (LLMs) augmented with retrieval have become a prominent approach to open-domain question answering and knowledge-intensive tasks. In a typical Retrieval-Augmented Generation (RAG) setup, the system searches a text corpus for relevant documents given a user query, then generates an answer grounded in the retrieved content \cite{lewis2020retrieval}. This grounding helps mitigate LLMs' tendency to hallucinate unsupported facts.
However, conventional RAG pipelines struggle when queries involve \textbf{dynamic or up-to-date information}. In such cases, static retrievers may return outdated or no relevant information, leading the LLM to either produce stale answers or hallucinate.
In this work, we present \textbf{DynaRAG}, a novel RAG framework that is \textit{dynamic-aware}: it can determine when static retrieval is insufficient and seamlessly incorporate external knowledge via API calls. DynaRAG first retrieves documents via web search, then reranks them with an LLM for relevance. If the top reranked score exceeds a threshold, the answer is generated from the top passages. Otherwise, DynaRAG automatically invokes an external API to fetch up-to-date information, which is merged with partial context for answer generation.
Our key contributions:
\begin{itemize}
    \item Propose DynaRAG, a RAG system with a dynamic decision mechanism for API fallback.
    \item Formalize the LLM-based reranking and sufficiency thresholding for triggering tool use.
    \item Integrate plug-in routing using FAISS-based schema filtering and Gorilla v2 \cite{patil2024gorilla} for robust API call generation.
    \item Demonstrate substantial improvements on dynamic and static QA benchmarks.
\end{itemize}
\section{Related Work}
\subsection{Retrieval-Augmented Generation}
RAG has become a standard approach for grounding LLMs in external knowledge without modifying their internal weights \cite{lewis2020retrieval}. However, most RAG pipelines rely on static sources like Wikipedia, and traditional benchmarks (e.g., Natural Questions, MS MARCO, QALD-10) focus on factual lookups in such frozen corpora \cite{nguyen2016ms}. This makes them ill-suited for evaluating performance on dynamic or time-sensitive queries.
To bridge this gap, the Comprehensive RAG (CRAG) benchmark \cite{yang2024crag} introduces tasks involving real-time web data and API access, along with a mix of popular and long-tail entities \cite{wu2024lora}. CRAG results reveal a significant performance drop in dynamic scenarios: GPT-4 \cite{achiam2023gpt} alone answers only $\sim$34\% of CRAG queries correctly, and a basic RAG pipeline improves this to just $\sim$44\%. These results underscore the limitations of static retrieval and motivate research into adaptive RAG systems that can better handle evolving knowledge.
\subsection{LLM-Based Reranking}
Retrieval quality plays a crucial role in RAG performance, prompting recent interest in using LLMs to rerank retrieved contexts. RankRAG \cite{yu2024rankrag} unifies reranking and answer generation by instruction-tuning a single model to perform both tasks. It blends ranking data into the instruction tuning phase, enabling the model to identify relevant passages before generating answers. This joint training approach surpasses traditional ranker pipelines while simplifying architecture. Inspired by this, we use an off-the-shelf LLM as a reranker at inference time.
\subsection{Tool-Augmented LLMs and API Usage}
LLMs can extend their capabilities through interaction with external tools and APIs. Systems like Gorilla \cite{patil2024gorilla} demonstrate how LLMs can map user intents to appropriate API functions. Gorilla, built on a fine-tuned LLaMA model, outperformed GPT-4 in producing correct API calls by leveraging large-scale API documentation. A key issue in tool-augmented LLMs is hallucination---models fabricating nonexistent APIs or parameters. Gorilla v2 addresses this via Retriever-Aware Training (RAT), where relevant API specs are retrieved at runtime and used to condition generation. This significantly reduces hallucinated calls and improves adaptability to evolving tool interfaces. Our method shares this retrieval-guided philosophy but applies it to natural language context rather than structured API specs.
\section{DynaRAG Methodology}
\subsection{System Overview}
\begin{figure}[htbp]
  \centerline{\includegraphics[width=0.7\columnwidth]{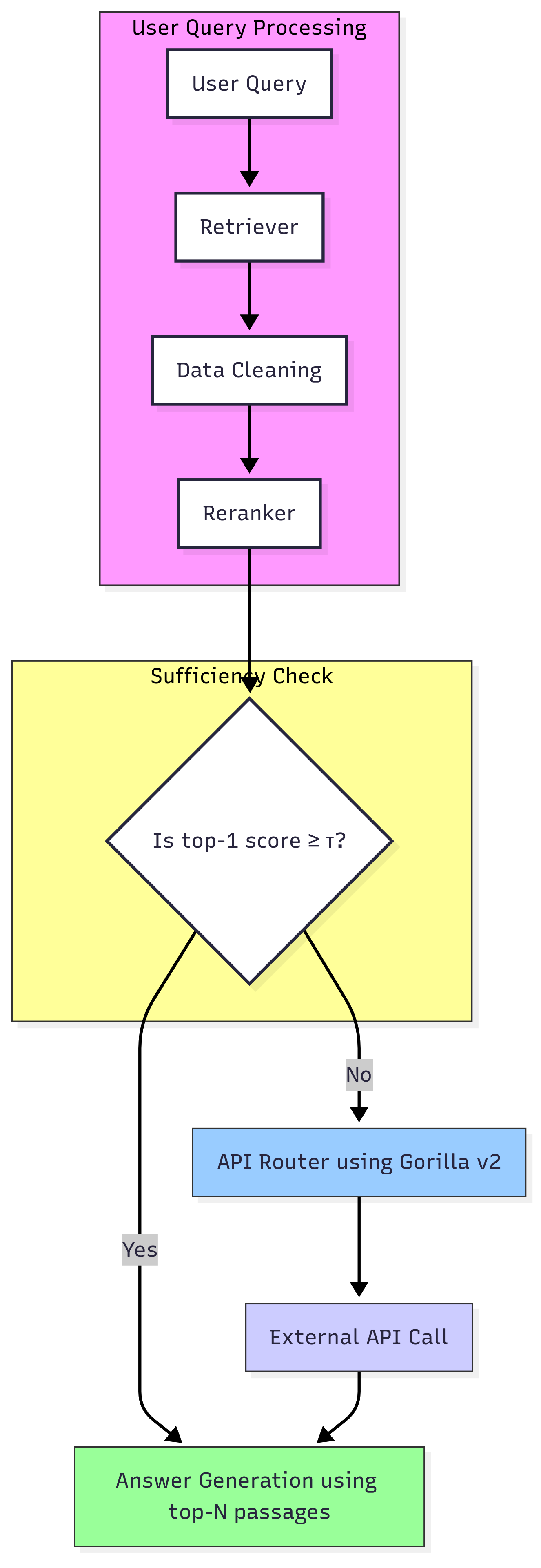}} % Adjust 0.8 to make it smaller/larger
  \caption{DynaRAG pipeline}
  \label{fig:architecture}
\end{figure}

DynaRAG consists of five main modules: \textbf{Retriever}, \textbf{Data Cleaning}, \textbf{Reranker}, \textbf{Sufficiency Classifier}, and \textbf{API Router}, followed by answer generation (Fig.~\ref{fig:architecture}).
\subsection{Retriever}
Given a user query $q$, we perform a web search (in our experiments, using the CRAG-provided pages) to obtain the top-$K$ raw HTML results:
\begin{equation}
    D_{\mathrm{ret}}(q) = \{d_{i_1}, \ldots, d_{i_K}\}\subset \mathcal{D}.
\end{equation}
\subsection{Data Cleaning}
Each retrieved document $d_i$ is an HTML page. Before passing to the reranker, we apply lightweight cleanup using \texttt{beautifulsoup4}:
\begin{itemize}
  \item Remove \texttt{<script>}, \texttt{<style>} and navigation extraneous tags.
  \item Extract and concatenate all paragraph (\texttt{<p>}) text.
  \item Normalize whitespace and strip headers/footers.
\end{itemize}
The result is a set of clean text snippets $\tilde{D}_{\mathrm{ret}}(q)=\{\tilde d_{i_1},\dots,\tilde d_{i_K}\}$.
\subsection{Reranking}
We feed each cleaned snippet $\tilde d_{i_j}$ along with the query into an LLM-based reranker:
\begin{equation}
    s_j = f_{\mathrm{rank}}\bigl(q,\,\tilde d_{i_j}\bigr).
\end{equation}
Sorting by $s_j$ yields the top-$N$ ranked passages $D_{\mathrm{rank}}(q)$.
\subsection{Sufficiency Classification}
We compare the highest reranking score $s_{(1)}$ against a threshold $\tau$:
\begin{equation}
    \delta(q) =
    \begin{cases}
      \text{``sufficient''} & \text{if } s_{(1)} \ge \tau,\\
      \text{``insufficient''} & \text{otherwise}.
\end{cases}
\end{equation}
If $\delta(q)=$\emph{``sufficient''}, the top $N$ passages are sent directly to the answer generator. Otherwise, we trigger the API fallback. We tune $\tau$ empirically on a held-out development set to maximize the sufficiency F1 score. While a fixed $\tau$ serves as an effective baseline for reducing unnecessary tool use, we note that optimal threshold values may vary across domains; future implementations could explore adaptive thresholding to better generalize to unseen query distributions.
\subsection{API Fallback with Gorilla v2}
When static retrieval is insufficient, we search our API schema catalog to retrieve the top-$M$ candidate schemas. These are passed to Gorilla v2, which---via Retriever-Aware Training---chooses and formats the correct API call. We execute the call, prepend its response to the context, and forward everything to the answer generator. This design decouples relevance assessment from tool invocation: the reranker acts as a high-precision filter, shielding the API router from irrelevant queries and reducing computational overhead. However, this dependency means that errors propagate downstream; a false ``sufficient'' classification prevents API access even if the static context is stale, highlighting the critical role of the sufficiency threshold.
\subsection{Answer Generation}
Finally, a generative LLM receives $q$, the top-$N$ reranked passages (if sufficient), and any API results (if fallback occurred), and produces the final, grounded answer.
\section{Experiments}
\subsection{Datasets and Metrics}
\subsubsection{CRAG Benchmark}
We adopt \textbf{CRAG} \cite{yang2024crag} as our primary benchmark because it uniquely captures the real-world challenges of RAG systems:
\begin{itemize}
  \item \emph{Realism:} Questions are drawn from smart-assistant use cases and paraphrased to reflect authentic user queries.
  \item \emph{Richness:} 4,409 QA pairs span five domains (Finance, Sports, Music, Movie, Open) and eight question types (simple, conditional, comparison, aggregation, multi-hop, set, post-processing, false-premise).
  \item \emph{Temporal Dynamism:} Facts cover four dynamism categories---real-time (seconds), fast-changing (daily), slow-changing (yearly), and static---allowing us to measure performance on truly time-sensitive queries.
  \item \emph{Popularity Diversity:} Entities range from head to long-tail, revealing how systems handle both common and rare knowledge.
\end{itemize}
These features make CRAG an ideal testbed for assessing how DynaRAG balances static retrieval with selective real-time API augmentation.
\subsection{Experimental Setup}
For \textbf{all} experiments we employ \textbf{Llama\,3--70B--Instruct} \cite{dubey2024llama} as the underlying language model. Hyper-parameters (temperature $=0$, top-p $=0.9$) are fixed across every setting to ensure comparability. Unless otherwise noted, we prepend the same concise system prompt:
\begin{quote}
\emph{``Answer in one or two short sentences. If you are not sure, respond with \texttt{I don't know}.''}
\end{quote}
This encourages brief answers and avoids speculative hallucinations.
\subsubsection{Evaluation Tasks}
Following \cite{yang2024crag}, we use two CRAG tasks:
\paragraph{Task 1} Up to \emph{five} web pages per question are supplied directly from Brave Search. The pages may be noisy and no structured knowledge-graph access is allowed; the task therefore isolates the answer-generation component of a RAG pipeline given a small, partially relevant unstructured context.
\paragraph{Task 2} In addition to the web context of Task 1, the system can query 38 mock APIs backed by a 2.6M-entity knowledge graph. Systems must decide which (if any) API to call, fuse the structured results with the unstructured snippets, and generate a grounded answer. Compared with Task 1, Task 2 stresses tool use, entity linking, and multi-source reasoning.
\subsubsection{Baselines}
We compare against two baselines:
\begin{itemize}
  \item \textbf{LLM Only:} The frozen Llama\,3--70B model prompted to answer directly, with no external context.
  \item \textbf{Direct RAG:} A baseline that appends a fixed-length web context window (4K tokens for Llama\,3--70B) to the prompt without reranking, sufficiency checking, or API routing. This simple strategy simulates minimal integration logic while exposing the model to potentially relevant web content.
\end{itemize}
\subsection{Evaluation}
\begin{table}[htbp]
\caption{Accuracy, hallucination, and missing rates for baseline and DynaRAG methods on the CRAG benchmark.}
\begin{center}
\renewcommand{\arraystretch}{1.2} % Added to improve readability and reduce density
\begin{tabular}{lccc}
\hline
 & \textbf{Acc.(\%)} & \textbf{Hall.(\%)} & \textbf{Miss.(\%)} \\
\hline
\textbf{LLM Only} & 28.53 & 34.95 & 36.52 \\
\textbf{Direct RAG} & 34.23 & 43.09 & 22.68 \\
\hline
\textbf{Task 1} & 29.12 & 25.33 & 45.55 \\
\textbf{Task 2} & 41.00 & 22.09 & 36.91 \\
\hline
\end{tabular}
\label{tab:crag-no-score}
\end{center}
\end{table}
Table~\ref{tab:crag-no-score} highlights how DynaRAG outperforms standard baselines on the CRAG benchmark, particularly under Task 2, which permits API augmentation. We observe several notable trends:
\paragraph{Improved Accuracy with API Fallback}
DynaRAG achieves 41.00\% accuracy on Task 2, a 6.77\% absolute gain over Task 1 (29.12\%), and substantially outperforms the LLM Only and Direct RAG baselines (28.53\% and 34.23\% respectively). This validates our hypothesis that selective API augmentation is critical for dynamic queries where static web documents fall short.
\paragraph{Reduced Hallucination}
Interestingly, hallucination rates decrease from 43.09\% in Direct RAG to 25.33\% in DynaRAG's Task 1 and further down to 22.09\% in Task 2. This suggests that LLM-based reranking and sufficiency filtering not only improve accuracy but also mitigate the model's tendency to fabricate unsupported content.
\paragraph{Missing Rate Trade-off}
While DynaRAG reduces hallucination, it incurs a higher missing rate in Task 1 (45.55\%) due to its conservative sufficiency thresholding mechanism. However, this is mitigated in Task 2 (36.91\%) through API fallback, which successfully retrieves relevant data when static context is lacking. This trade-off exemplifies the benefit of dynamic routing: when the system is unsure, it either abstains or seeks additional structured evidence, increasing overall truthfulness.
\paragraph{Effectiveness of Dynamic Awareness}
Compared to Direct RAG, which naively appends retrieved documents, DynaRAG's dynamic-aware control flow yields more robust performance. The integration of a sufficiency classifier enables precise decisions on when to trust retrieved text versus when to fall back on APIs. These results collectively demonstrate that DynaRAG's design---combining LLM reranking, sufficiency thresholding, and Gorilla-powered API generation---effectively bridges the gap between static and dynamic knowledge sources, leading to more accurate and reliable answers in time-sensitive QA settings.
\section{Conclusion}
We presented \textbf{DynaRAG}, a dynamic-aware Retrieval-Augmented Generation (RAG) framework that combines LLM-based reranking, sufficiency classification, and Gorilla v2-driven API fallback to address both static and time-sensitive queries. Our system dynamically decides whether retrieved documents are sufficient or whether an external API should be invoked, bridging the gap between frozen knowledge and evolving real-world information. Experiments on the CRAG benchmark demonstrate that DynaRAG significantly improves accuracy on dynamic questions, reduces hallucinations compared to baseline RAG systems, and maintains strong performance on static tasks. These results underscore the importance of integrating selective, context-aware tool use in modern QA systems. DynaRAG offers a promising step toward building LLM assistants that are not only grounded and reliable, but also capable of real-time knowledge integration in a rapidly changing world.
\section{Limitations}
While our dynamic-aware RAG framework demonstrates significant improvements, several limitations warrant discussion.

\textbf{API Catalog Dependency.} DynaRAG depends on a prebuilt set of structured API schemas. While Gorilla v2 generalizes well to unseen APIs, the system's effectiveness is bounded by the availability and quality of API documentation. In domains with sparse or outdated API specs, fallback performance may degrade\cite{xue2026resilientroutingriskawaredynamic}.

\textbf{Latency and Computational Overhead.} Incorporating reranking, sufficiency classification, and real-time API invocation introduces nontrivial latency and resource demands. Although beneficial for accuracy and hallucination reduction, this added complexity may limit DynaRAG's use in low-latency or cost-sensitive environments.

\textbf{Static Sufficiency Thresholding.} The decision to trigger API fallback is based on a manually selected confidence threshold from the reranker. This fixed setting may not adapt well across different query distributions or domains, potentially leading to suboptimal routing decisions.

\textbf{Benchmark Scope.} Our evaluation is currently centered on the CRAG benchmark. While CRAG covers a wide range of domains and dynamism levels, the generalizability of DynaRAG to other benchmarks or real-world QA settings---especially those involving long-tail entities or rapidly evolving knowledge bases---requires further exploration. Future work will extend our validation to additional dynamic knowledge benchmarks, such as RealTimeQA \cite{kasai2023realtime}, to fully stress-test the API router and sufficiency classifier across diverse, time-sensitive datasets.\bibliographystyle{IEEEtran}
\bibliography{custom}
\end{document}